\documentclass[runningheads]{llncs}
\usepackage[T1]{fontenc}
\usepackage{graphicx}
\usepackage{hyperref}
\usepackage{xcolor}
\usepackage{bm}
\usepackage{dsfont}
\usepackage{booktabs}
\usepackage{multirow}
\usepackage{amsmath}
\usepackage{comment}
\usepackage[noadjust]{cite}
\def\discintname{Disclosure of Interests.}

\begin{document}
\title{Understanding the Dependence of Perception Model Competency on Regions in an Image}
\titlerunning{Understanding Perception Model Competency}
\author{Sara Pohland\orcidID{0000-0003-2746-6372} \and
Claire Tomlin\orcidID{0000-0003-3192-3185}}
\authorrunning{S. Pohland \& C. Tomlin}
\institute{University of California, Berkeley CA 94720, USA\\
\email{\{spohland,tomlin\}@berkeley.edu}}
\maketitle
\begin{abstract}
While deep neural network (DNN)-based perception models are useful for many applications, these models are black boxes and their outputs are not yet well understood. To confidently enable a real-world, decision-making system to utilize such a perception model without human intervention, we must enable the system to reason about the perception model’s level of competency and respond appropriately when the model is incompetent. In order for the system to make an intelligent decision about the appropriate action when the model is incompetent, it would be useful for the system to understand \textit{why} the model is incompetent. We explore five novel methods for identifying regions in the input image contributing to low model competency, which we refer to as image cropping, segment masking, pixel perturbation, competency gradients, and reconstruction loss. We assess the ability of these five methods to identify unfamiliar objects, recognize regions associated with unseen classes, and identify unexplored areas in an environment. We find that the competency gradients and reconstruction loss methods show great promise in identifying regions associated with low model competency, particularly when aspects of the image that are unfamiliar to the perception model are causing this reduction in competency. Both of these methods boast low computation times and high levels of accuracy in detecting image regions that are unfamiliar to the model, allowing them to provide potential utility in decision-making pipelines \footnote{The code for reproducing our methods and results is available on GitHub: https://github.com/sarapohland/explainable-competency.}.

\end{abstract}

\section{Importance of Understanding Model Competency}



Deep neural networks (DNNs) have been very successful in image classification tasks. However, DNN models are considered black boxes, and there currently exists only a shallow understanding of how these models obtain their outputs \cite{rauker_toward_2023}. This greatly limits our ability to trust these models in real-world systems, where their failure could be detrimental without human intervention. To confidently employ DNN-based perception models in the real world, we need to develop more robust and generalizable models, but we also need to gain a deeper understanding of how these models work and when they might fail. There has been much work on quantifying uncertainty in perception systems \cite{gawlikowski_survey_2023}, as well as identifying when an input is out-of-distribution (OOD) \cite{yang_generalized_2022}. However, with only knowledge that the model is uncertain or that an input is OOD, the systems that utilize these perception models are limited in their ability to appropriately respond when the perception model lacks confidence. It would be useful to understand \textit{why} a model is believed to be uncertain and \textit{why} an input is deemed to be OOD before the output of a perception model is used for decision-making. 

We explore these ideas in this paper, developing methods to identify particular regions in an image that lead to a lack of model confidence. Towards this end, we explore five approaches for identifying regions in an image contributing to low model competency. We begin by considering a naive approach, in which images are partitioned into regions and the competency associated with each cropped region is assessed. We then explore segment masking, pixel perturbation, and competency gradient approaches that are related to methods present in artificial intelligence (AI) explainability work. Finally, we present a novel method for understanding regions contributing to model incompetency that is based on autoencoder reconstruction loss. We develop methods for evaluating each of these approaches and compare their performance across three diverse datasets. We present an analysis of the results and conclude with avenues for future work.

\section{Background \& Related Work}

In this work, we explore explanations for \textit{model competency}--a generalized form of predictive uncertainty. Predictive uncertainty generally arises from three factors: data (or aleatoric) uncertainty, model (or epistemic) uncertainty, and distributional shift \cite{alice}. Data uncertainty refers to uncertainty arising from complexities of the data (i.e., noise, class overlap, etc.), while model uncertainty reflects the ability of the perception model to capture the true underlying model of the data \cite{yarin_gal_uncertainty_2016}. These types of uncertainty are explored in work referred to as uncertainty quantification (Section \ref{uncertainty-quant}). Distributional shift refers to mismatched training and test distributions \cite{quinonero-candela_dataset_2009}. Extensive work has explored methods to detect inputs that are out-of-distribution (OOD) (Section \ref{ood}). This paper aims to expand upon work on uncertainty quantification and OOD detection with the goal of offering explanations for why a perception model's prediction is uncertain. This idea of explaining perception model competency is related to previous work within the area of explainable image classification (Section \ref{explainable}), which seeks to offer explanations for the classification decisions of perception models. Our work differs from existing work in this area in that we seek to offer explanations for low model competency, rather than explanations for a model's prediction.

\subsection{Uncertainty Quantification} \label{uncertainty-quant}

Extensive research has been done on methods to understand and quantify uncertainty in a neural network's prediction. The modeling of these uncertainties can generally be divided into methods based on (1) deterministic neural networks, (2) Bayesian neural networks (BNNs), and (3) ensembles of neural networks \cite{gawlikowski_survey_2023}. Approaches that use a single deterministic neural network estimate model uncertainty based on a single forward pass within a deterministic network. These approaches are often either distance-based \cite{ramalho_density_2019,liu_simple_2020} or gradient-based \cite{oberdiek_classification_2018,lee_gradients_2020}, although many other approaches exist \cite{gawlikowski_survey_2023}. A BNN is a stochastic model whose output is a probability distribution over its predictions \cite{neal-1992,neal_bayesian_1996}. Approaches that employ BNNs extract uncertainty as a statistical measure, often using sampling methods \cite{welling-2011}, variational inference \cite{graves-2011,rezende-2015,gal_dropout_2016}, or Laplace approximation \cite{ritter2018a}. Ensemble methods combine the predictions of multiple deterministic networks to form a probability density function \cite{lakshminarayanan_simple_2017}. While all of these methods provide an estimate of the uncertainty associated with a model's prediction, none of these offer rationale for \textit{why} the model is uncertain.

\subsection{Out-of-Distribution (OOD) Detection} \label{ood}

Many recent approaches have focused on quantifying distributional uncertainty--uncertainty that is caused by a change in the input data distribution \cite{quinonero-candela_dataset_2009}. Approaches that are specifically focused on determining if an input falls outside of the input-data distribution are referred to as OOD detection methods \cite{yang_generalized_2022}. These approaches are generally either (1) classification-based, (2) density-based, (3) distance-based, or (4) reconstruction-based. Classification-based methods is a broad category that captures methods that largely rely on classifier predictions. These methods generally seek to revise the overly confident softmax scores at the output of neural networks to detect OOD samples more robustly \cite{liang_enhancing_2020,hsu_generalized_2020,liu_energy-based_2020}. Density-based methods model the training distribution with some probabilistic model and flag test data in low-density regions as OOD. These approaches tend to rely on Gaussian mixture models (GMMs) \cite{zong2018deep}, kernel density estimation (KDE) \cite{kde}, normalizing flows \cite{rezende_variational_2016,kingma_glow_2018}, or likelihood ratios \cite{ren-2019}. Distance-based methods use distance metrics in the feature space with the assumption that OOD samples should lie relatively far from the centroids or prototypes of the training classes. Many popular approaches utilize the Mahalanobis distance \cite{lee-2018,Zaeemzadeh_2021_CVPR}, cosine similarity \cite{Techapanurak_2020_ACCV}, or a nearest-neighbor distance \cite{sun_out--distribution_2022}. Finally, reconstruction-based methods rely on the reconstruction loss of autoencoders (AEs) \cite{xia-2015,gong_memorizing_2019}, variational autoencoders (VAEs) \cite{An2015VariationalAB}, convolutional autoencoders (CAEs) \cite{chen-2018}, or generative adversarial networks (GANs) \cite{zenati_efficient_2019,Sabokrou,OCGAN}. While all these methods have been used to successfully identify OOD samples, they do not offer a clear explanation as to \textit{why} the samples are deemed OOD.


\subsection{Explainable Image Classification} \label{explainable}

The field of explainable artificial intelligence (XAI) seeks to offer explanations for why a model makes the decisions that it does with the goal of developing more interpretable and trustworthy models \cite{rauker_toward_2023,ali_explainable_2023}. Within the area of explainable image classification, there exist (1) posthoc techniques, which explain the behavior of an AI system after it has been trained without modifying the underlying model architecture, and (2) antehoc methods, which incorporate the aspect of explainability into the model design process to generate classification models that are inherently more interpretable \cite{kamakshi_explainable_2023}. Our work relates more closely to posthoc techniques, which employ various methods to understand the workings of black-box models and explain their decision-making process. 

The most common posthoc approach for explaining an image classifier is to identify key image regions contributing to the model’s predictions and display them using a saliency map. These methods display the regions salient toward the prediction of a class through gradient-based approaches \cite{simonyan_deep_2014,grad-cam,Sundararajan}, perturbation-based approaches \cite{Fong_2017_ICCV,zintgraf2017visualizing,score-cam}, layerwise-relevance propagation \cite{bach_pixel-wise_2015,montavon_explaining_2017,Shrikumar}, masking-based approaches \cite{Dabkowski}, or deconvolution methods \cite{fleet_visualizing_2014}. Other approaches aim to construct an inherently interpretable pseudo-classifier that approximates the working mechanism of the black box classifier, offering either local interpretations around a particular instance \cite{Ribeiro,ribeiro_anchors_2018} or global interpretations \cite{Lundberg,lundberg_consistent_2019}. There are also approaches that provide explanations based on abstract vector representations, which can be linked to human-understandable concepts. These methods tend to use concept bottleneck models \cite{koh-2020} or concept activation vectors \cite{kim_interpretability_2018} if the categories of interest are provided, or they focus on discovering task-relevant concepts automatically \cite{Ghorbani,yeh-2020}. Finally, there are methods that offer counterfactual explanations by generating alternative scenarios to explain the behavior of the classifier \cite{Goyal}. While all of these methods enhance the understanding of image classifiers, none of them explicitly deals with model \textit{competency}. These methods do not offer an estimate of or explanation for how confident a model is in its prediction (as we demonstrate in Appendix \ref{cams}).

\subsection{Explainable Competency Estimation}

While there has been work on quantifying model uncertainty, identifying OOD inputs, and explaining model predictions, to the best of our knowledge, there has not been an effort to offer explanations for the lack of competency associated with a model's prediction. We seek to bridge this gap by bringing together work on uncertainty quantification, OOD detection, and explainability. We see this as an important step in developing decision-making systems that can reason about model competency and respond safely in the face of uncertainty. 

\section{Approach for Understanding Model Competency}


Recall that \textit{model competency} is a generalized form of predictive uncertainty that captures data (or aleatoric) uncertainty, model (or epistemic) uncertainty, and distributional shift \cite{alice}. To gain a deeper understanding of perception model competency, we explore five methods for visualizing regions in the input image that contribute most to a lack of competency (similar to saliency methods discussed in Section \ref{explainable}). Towards this end, we first compute a competency score, estimating the confidence in a model's prediction for a given input image (Section \ref{est-comp}). If the model is deemed incompetent (i.e., the competency score is below some threshold), we seek to understand which regions in the image most lead to this lack of competency by estimating the dependence of the competency score on segmented regions in the image (Section \ref{seg-comp}).

\subsection{Estimating Model Competency} \label{est-comp}

Let $f$ be the true underlying model from which our images are drawn and $\hat{f}$ be the predicted model (or the perception model). For an input image, $\bm{X}$, the perception model aims to estimate the true class of the image, $f(\bm{X})$, from the set of all classes. The competency of the model for this image is given by
\begin{equation}
  C(\bm{X}) := P\bigl(\{\hat{f}(\bm{X})=f(\bm{X})\}|\bm{X}\bigr).  
\end{equation}

To simplify our notation, let $\hat{c}$ be the class predicted by the deterministic perception model (i.e., $\hat{f}(\bm{X})=\hat{c}$), so we can express the model competency as
\begin{equation}
    C(\bm{X}) = P\bigl(\{f(\bm{X})=\hat{c}\}|\bm{X}\bigr).
\end{equation}

It is difficult to estimate this probability directly because we are limited by the data contained in the training sample. Let $D$ be the event that the input image is in-distribution (i.e., drawn from the same distribution as the training samples) and write the following lower bound on competency:
\begin{align}
    C(\bm{X}) &\geq P\bigl(D \cap \{f(\bm{X})=\hat{c})\}|\bm{X}\bigr) \\
    &= P\bigl(\{f(\bm{X})=\hat{c}\}|\bm{X}, D\bigr)P(D|\bm{X}).
\end{align}

These probabilities are not readily available but can be approximated \cite{alice}. To estimate the first probability, we can fit a calibrated transfer classifier, such as the logistic regression classifier, and use its class probability output to obtain an estimate, $\hat{p}_{\hat{c}|D}(\bm{X})$. To estimate the probability that an image is from a class distribution, we can model the training distribution as a Gaussian mixture model (GMM) and compute the Mahalanobis distance of the feature vector to each of the class distributions. We then use these distances as input to another logistic regression classifier to estimate the probability that the input image came from the training distribution, $\hat{p}_D(\bm{X})$. Using these approximations, the lower bound on the probability of model competency can be approximated as
\begin{equation}
    C(\bm{X}) \geq \hat{p}_{\hat{c}|D}(\bm{X})\hat{p}_D(\bm{X}).
\end{equation}


We refer to this expression as the \textit{competency score}. This metric ties together work on uncertainty estimation (Section \ref{uncertainty-quant}) and OOD detection (Section \ref{ood}), using a deterministic classification-based approach to estimate model uncertainty assuming a sample is in-distribution and a distance-based approach to estimate the probability that a sample is in-distribution. Note that the competency score estimates a probability and is thus always between zero and one. If the model is 100\% competent on a given image, the probability that the model's prediction is correct is one. If the model is entirely incompetent, the probability that the prediction is correct is zero. The competency score is thus between zero and one, with higher scores corresponding to higher levels of competency. 
Another important aspect of this metric, that distinguishes is from many existing uncertainty and OOD estimates, is that it is differentiable with respect to the input image.

\subsection{Identifying Regions Contributing to Low Competency} \label{seg-comp}

We explore five approaches for estimating the dependence of the competency score on regions in the input image to understand the extent to which different regions in the image contribute to low model competency. 


\subsubsection{Approach 1: Image Cropping} 
\begin{figure}[h!]
    \centering
    \includegraphics[width=0.9\columnwidth]{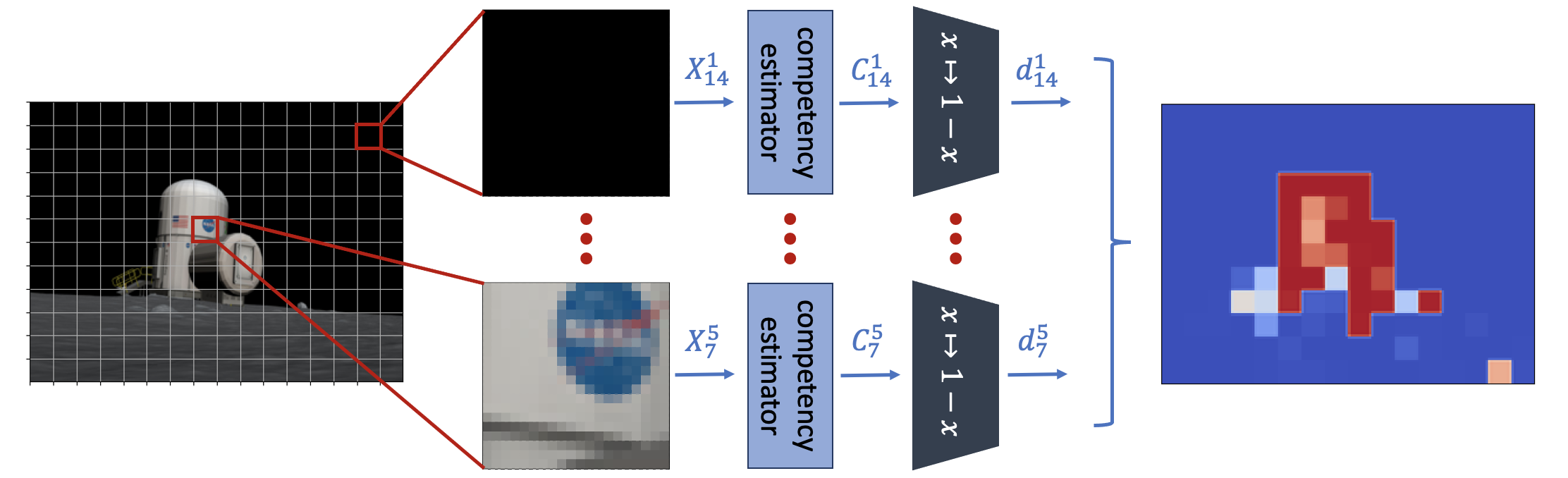}
    \caption{Image cropping approach for identifying low competency regions. We partition the image into grid cells, crop the image around each grid cell to obtain a new image, $X^i_j$, and compute the competency score, $C^i_j$, for each cropped image. Regions with lower scores are associated with lower levels of competency and are said to contribute more to the overall low model competency, resulting in a higher dependency score, $d^i_j$.}
    \vspace{-10mm}
    \label{fig:cropping}
\end{figure}

The first naive approach we consider is referred to as \textit{Cropping} and is described in Figure \ref{fig:cropping}. For this approach, we simply compute the competency score for each region in the image within a grid. We tune the height and width of the grid cells to achieve the best performance.


\subsubsection{Approach 2: Segment Masking} 
\begin{figure}[h!]
    \centering
    \includegraphics[width=0.9\columnwidth]{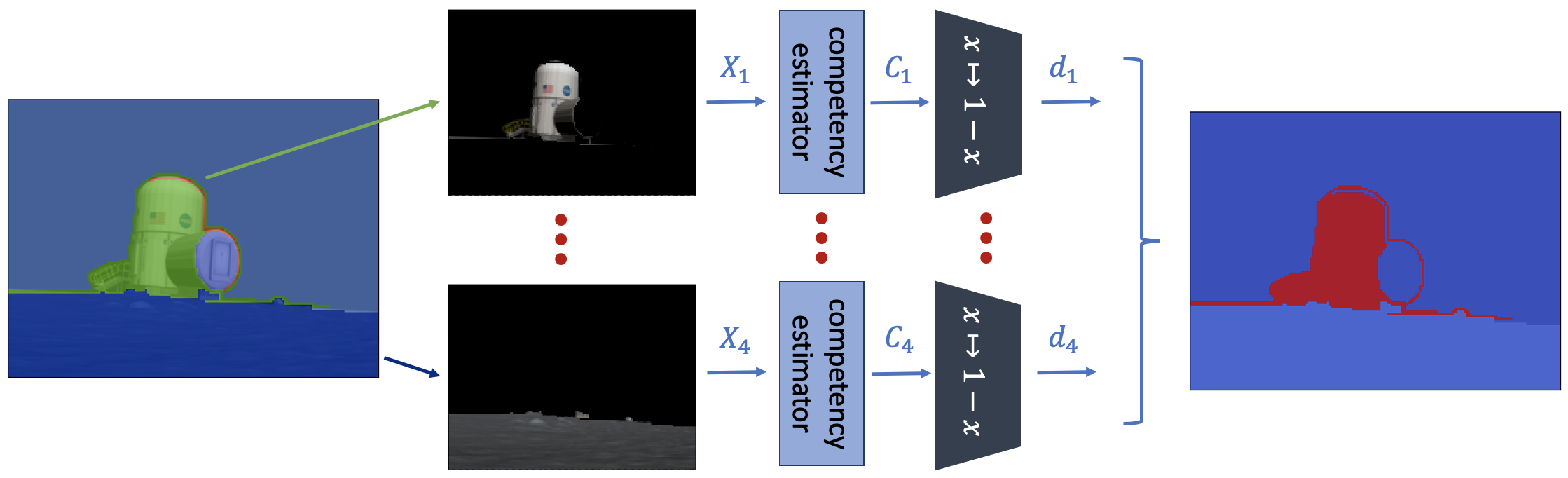}
    \caption{Segment masking approach for identifying low competency regions. We begin by segmenting the image using the Felzenszwalb segmentation algorithm \cite{felzenszwalb_efficient_2004}. For each segment determined by this algorithm, we mask out the rest of the image to obtain a new image, $X_i$, and compute the competency score, $C_i$, of that masked image. Segments with lower corresponding competency scores are said to contribute more to the overall low model competency for that image, resulting in a higher dependency score, $d_i$.}
    \vspace{-5mm}
    \label{fig:masking}
\end{figure}

We refer to the second approach considered as \textit{Masking}. In this approach, we compute the competency score for each segment in the image, while masking out the rest of the image, as shown in Figure \ref{fig:masking}. We explore various methods for masking in an attempt to capture ``feature missingness" \cite{sturmfels_visualizing_2020}--blurring regions that are not of interest, adding noise to those regions, or replacing those pixels with zeros, ones, average values, uniform random values, or Gaussian random values--and select the highest-performing method. 


\subsubsection{Approach 3: Pixel Perturbation} 
\begin{figure}[h!]
    \centering
    \includegraphics[width=0.9\columnwidth]{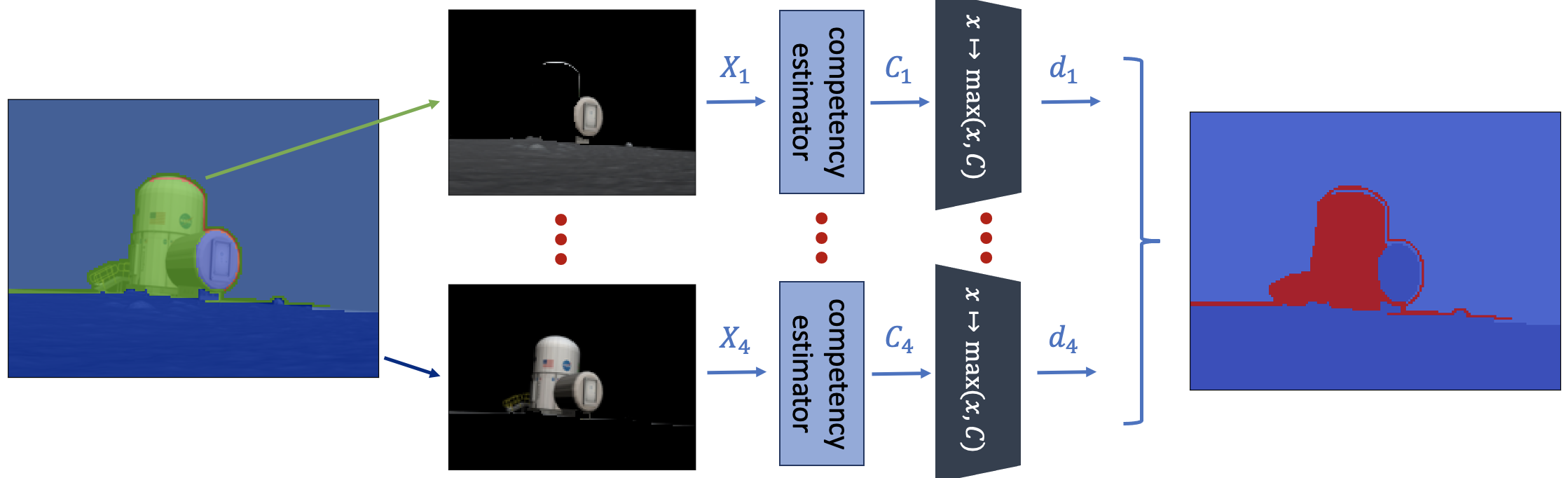}
    \caption{Pixel perturbation approach for identifying low competency regions. We begin by segmenting the image using the Felzenszwalb algorithm \cite{felzenszwalb_efficient_2004}. We then perturb each segment determined by this algorithm to obtain a new image, $X_i$, and compute the competency score, $C_i$, for each of these perturbed images. If the competency score increases (above $C$) after perturbing a given region, that region is believed to contribute more to the overall low model competency, resulting in a higher dependency score, $d_i$.}
    \vspace{-10mm}
    \label{fig:perturbation}
\end{figure}

The third approach, which is closely related to the second, is referred to as \textit{Perturbation} and is described in Figure \ref{fig:perturbation}. In this approach, we measure the increase in competency achieved by perturbing segments of the image. This approach is related to perturbation approaches in XAI that predict the importance of pixels on a model's prediction by perturbing those pixels and observing the resulting change in output \cite{Fong_2017_ICCV}. We consider various methods for perturbing regions of interest, as discussed for \textit{Masking}. 

\subsubsection{Approach 4: Competency Gradients} 
\begin{figure}[h!]
    \centering
    \includegraphics[width=0.8\columnwidth]{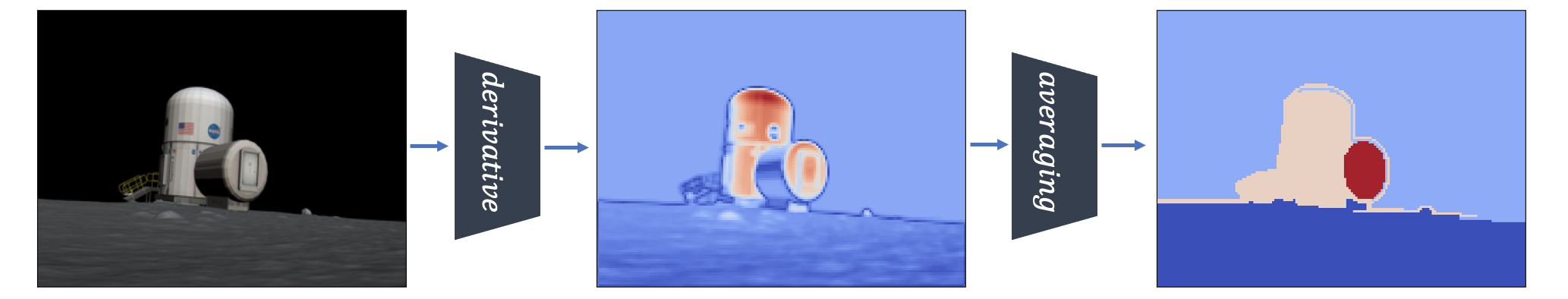}
    \caption{Competency gradients approach for identifying low competency regions. We begin by computing the partial derivative of the overall competency score with respect to each of the pixel values in the input image. We then calculate the average derivative over segmented regions in the image, using the Felzenszwalb algorithm \cite{felzenszwalb_efficient_2004}.}
    \vspace{-10mm}
    \label{fig:gradients}
\end{figure}

The fourth approach, which we refer to as \textit{Gradients}, relies on gradient information relating the input image to the estimated competency score and is described in Figure \ref{fig:gradients}. The competency score (defined as $C$ in Section \ref{est-comp}) is a differentiable function of the input image, $\bm{X}$. Therefore, we can compute the partial derivative of the competency score with respect to each of the pixel values in the input image. If the derivative is large, then small changes in this pixel value will significantly affect the competency score, and we say that model competency is more dependent on this pixel. This approach is similar to gradient-based approaches in XAI \cite{grad-cam} but considers the dependence of the competency score on segmented regions, rather than the dependence of the model output on individual pixels. 


\subsubsection{Approach 5: Reconstruction Loss} 
\begin{figure}[h!]
    \centering
    \includegraphics[width=0.9\columnwidth]{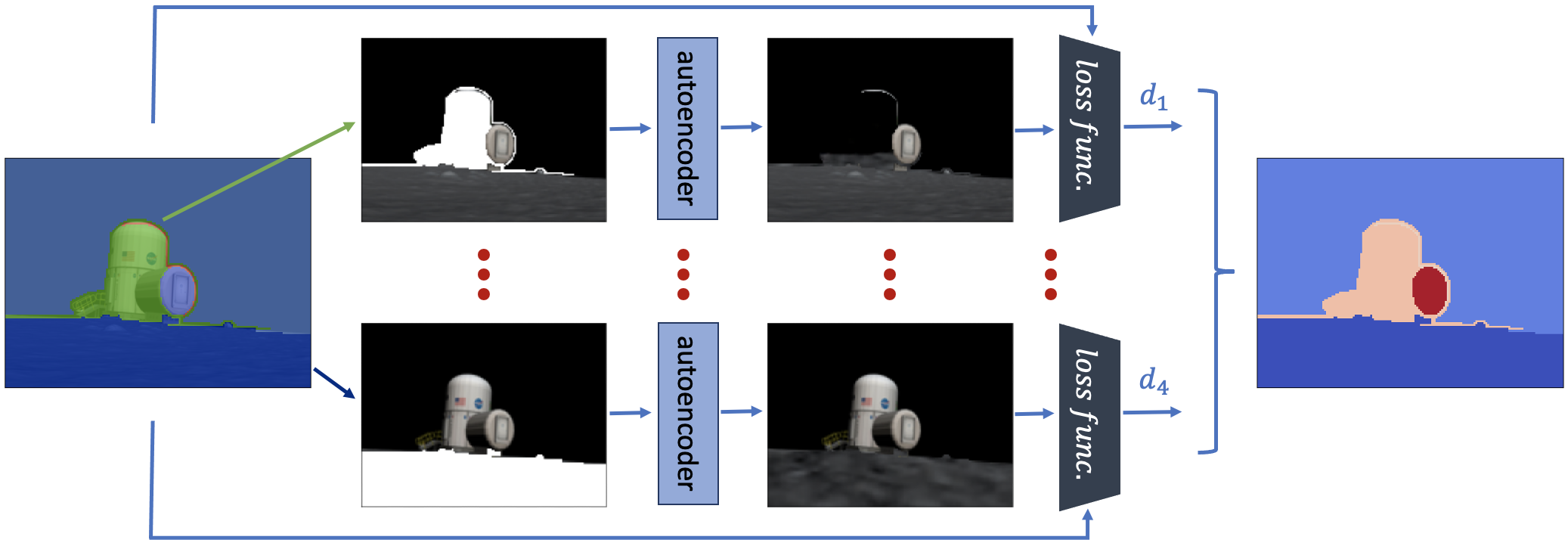}
    \caption{Reconstruction loss approach for identifying low competency regions. We begin by segmenting the image using the Felzenszwalb algorithm \cite{felzenszwalb_efficient_2004}, then generate masked images for each segment. Given an image with a segment masked out using ones, an autoencoder aims to predict the pixels of the original image. The difference between the original image and the predicted image is the reconstruction loss. If the reconstruction loss is high for a given image segment, this segment is believed to be unfamiliar to the perception model, and the overall low model competency is said to be more dependent on this region in the image, resulting in a higher dependency score, $d_i$.}
    \label{fig:reconstruction}
\end{figure}

The fifth and final approach we considered is referred to as \textit{Reconstruction} because it depends on the reconstruction loss of an autoencoder trained to fill in missing regions of an input image, as shown in Figure \ref{fig:reconstruction}. The autoencoder is designed to reconstruct the original image from the feature vector used for prediction by the perception model. We assume that the ability of the autoencoder to accurately reconstruct a region of an image reflects the familiarity of the model with that image region and thus the dependence of low model competency on that region. Reconstruction-based approaches have been explored for anomaly detection \cite{gong_memorizing_2019} and localization \cite{zavrtanik_reconstruction_2021} but have not yet been explored as an explainability tool. For determining the reconstruction loss across segmented regions in an image, we considered different methods and loss functions and selected those that achieved the best performance.


\section{Method Evaluation \& Analysis}

We provide a thorough analysis of the performance and usefulness of each of these five approaches for identifying regions contributing to low model competency. We conduct analysis across three unique datasets that assess the ability to identify unfamiliar objects (Section \ref{lunar}), detect regions associated with unseen classes (Section \ref{signs}), and recognize unexplored areas in an environment (Section \ref{park}).

\subsection{Metrics for Comparison}

Following much of the work in explainable image classification (Section \ref{explainable}), we provide a \textit{visual} comparison of the outputs of the five approaches. This visual analysis allows us to better understand the method outputs and determine the extent to which we trust each of the methods for explaining model competency. It also helps to determine whether the outputs of these methods offer explanations that are helpful to and interpretable by a human user. While we cannot display every example image in this paper, we include a few representative images.

While visual analysis is useful, it may be misleading to rely solely on a qualitative comparison \cite{sturmfels_visualizing_2020}. In an effort to provide a more rigorous quantitative comparison, we develop a metric for \textit{accuracy} that evaluates the ability of each of the five approaches to identify regions that are unfamiliar to the model. Each image in the test set is manually labeled to indicate which regions contain features that were not present in the training set. A pixel that corresponds to a feature that was not present in the training set is considered a positive sample, and a pixel that corresponds to a feature that was present is considered a negative sample. We then assess the ability of the five approaches to label the dataset in the same way, measuring the overall accuracy, true positive rate (TPR), true negative rate (TNR), positive predictive value (PPV), and negative predictive value (NPV) averaged across all of the images in the test set. 

Finally, in order for any of these approaches to truly be useful for decision-making, we must consider the \textit{computation time} required to identify regions contributing to low competency. While it is valuable to provide users a better understanding of how their model works, when it is incompetent, and what leads to incompetency, it is desirable for systems to recognize their level of competency and respond appropriately without the need for human intervention. In order for this autonomy to be realistic, the system needs to reason about its competency fast enough that doing so does not interrupt the decision-making process.

\subsection{Dataset 1: Lunar Environment} \label{lunar}

\begin{figure}[h]
    \centering
    \includegraphics[width=\columnwidth]{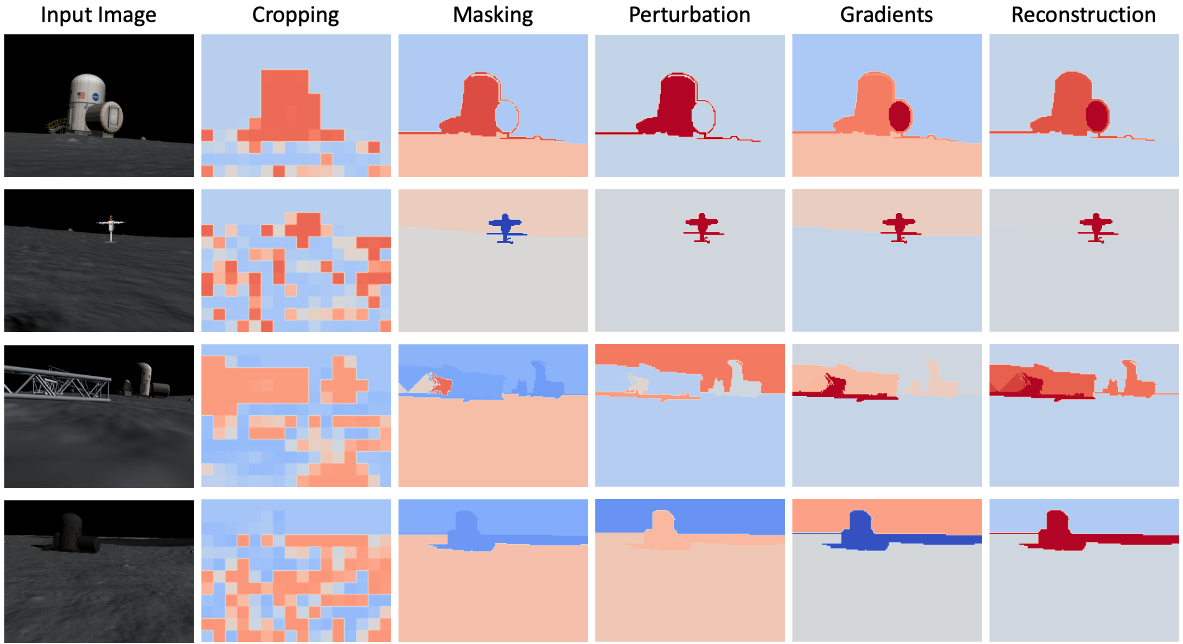}
    \caption{Visual comparison of regional competency explanation methods for a lunar environment. Rows correspond to different input images and columns to different techniques. More red regions are more associated with low levels of model competency.}
    \label{fig:lunar}
    \vspace{-5mm}
\end{figure}

The first dataset we consider was obtained from a lunar environment developed by the Johns Hopkins University Applied Physics Laboratory in the Gazebo simulator, in which the training data contains images from an uninhabited moon and the test data contains images from a habited moon. While the training images only contain the lunar surface, the sky, and shadows, the test images additionally contain astronauts and human-made structures. This dataset assesses the ability of the regional competency approaches discussed in Section \ref{seg-comp} to identify unfamiliar objects. A visual comparison of these approaches is shown in Figure \ref{fig:lunar} and a quantitative comparison is given in Table \ref{tab:lunar}. We also consider the performance of existing class activation mapping techniques for identifying unfamiliar objects in Figure \ref{fig:cam-lunar} and Table \ref{tab:cam-lunar} of Appendix \ref{cams-lunar}.

Upon visual inspection, it is apparent that the \textit{Cropping} method can roughly identify regions in the lunar images with unfamiliar objects, but this method is quite noisy, limiting its utility. This method has the worst overall accuracy, as it is quite conservative and often classifies regions that should be familiar to the model as ones leading to low model competency. In addition, this method is the slowest by far and may be too slow for many decision-making problems.

The four approaches that use segmentation appear more useful for identifying unfamiliar objects. The \textit{Masking} method does well on some examples in the lunar dataset, but it also often inverts the expected order of dependency, giving low dependency scores to many regions that contain unfamiliar objects. The outputs of the \textit{Perturbation} method more often align with what is expected, but this method also fails to identify several unfamiliar objects. The \textit{Gradients} method tends to perform quite well when the unfamiliar structures look significantly different from elements of the uninhabited moon, but its performance deteriorates when the image contains dark structures that blend into the background. 

These three methods perform comparably in terms of overall accuracy, but the \textit{Gradients} method more often identifies regions as containing unfamiliar features, resulting in a higher TPR and lower TNR compared to the \textit{Masking} and \textit{Perturbation} methods. Visually, the \textit{Reconstruction} method appears to be the most useful and reliable; it consistently assigns high scores to unfamiliar structures and generally always seems to outperform the other methods. Quantitatively, it also distinguishes between familiar and unfamiliar features most accurately. This method achieves the highest overall accuracy, along with the highest TPR, TNR, PPV, and NPV. In terms of computation time, the \textit{Gradients} method is fastest, but the \textit{Masking}, \textit{Perturbation}, and \textit{Reconstruction} methods are also reasonably fast, obtaining their outputs in around 0.15 seconds or less.

\begin{table}[h]
\centering
\vspace{-2mm}
\caption{Quantitative comparison of regional competency explanation methods for a lunar environment. A true positive occurs when a method correctly identifies a pixel as one corresponding to an unfamiliar object (i.e., an astronaut or human-made structure), and a true negative occurs when a method identifies a pixel as one corresponding to a feature present in the training set (i.e., the sky, the lunar surface, or a shadow).}
\begin{tabular}{|c|c|c|c|c|c|c|}
\hline 
\multirow{2}{*}{Method} & \multirow{2}{*}{\shortstack{Average \\ Time (sec)}} & \multicolumn{5}{c|}{Average Accuracy (\%)} \\
\cline{3-7}
& & \hspace{2mm}Overall\hspace{2mm} & \hspace{2mm}TPR\hspace{2mm} & \hspace{2mm}TNR\hspace{2mm} & \hspace{2mm}PPV\hspace{2mm} & \hspace{2mm}NPV\hspace{2mm} \\
\hline 
Cropping & 1.539 & 71.27 & 70.01 & 72.97 & 26.48 & 92.03 \\
Masking & 0.151 & 84.01 & 33.07 & 92.99 & 65.06 & 87.62 \\
Perturbation & 0.116 & 83.53 & 29.74 & 92.57 & 59.22 & 87.02 \\
Gradients & \textbf{0.070} & 82.16 & 51.32 & 87.59 & 64.44 & 91.01 \\
Reconstruction & 0.081 & \textbf{96.16} & \textbf{88.87} & \textbf{98.83} & \textbf{93.53} & \textbf{95.95} \\
\hline
\end{tabular}
\vspace{-5mm}
\label{tab:lunar}
\end{table}

\subsection{Dataset 2: Speed Limit Signs} \label{signs}

\begin{figure}[h]
    \centering
    \includegraphics[width=\columnwidth]{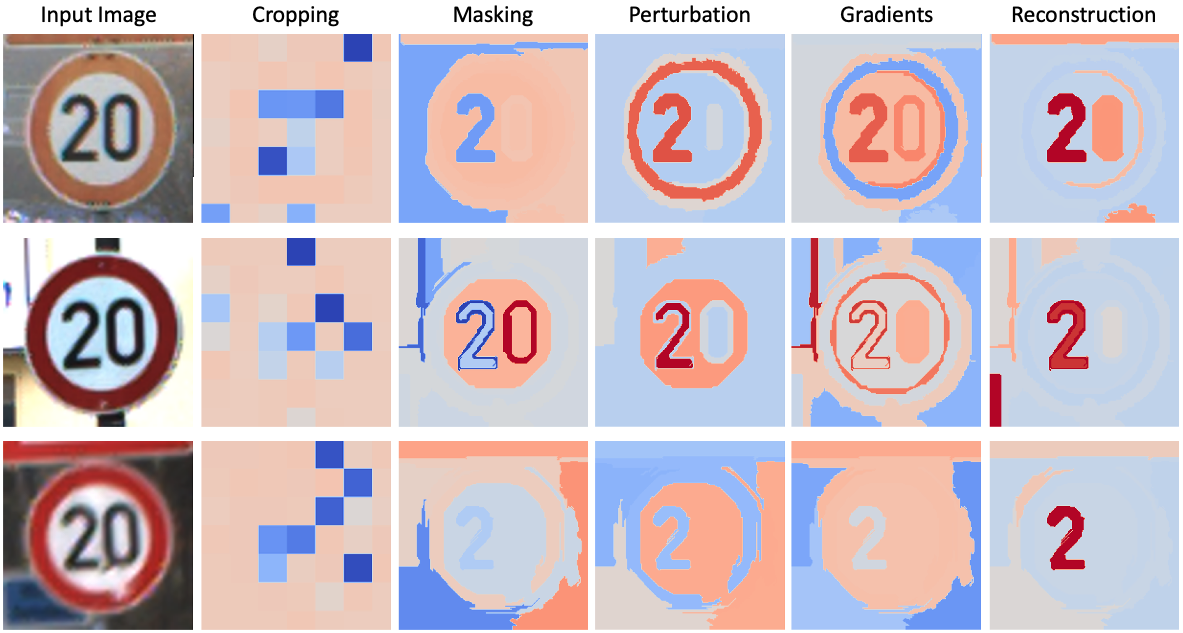}
    \caption{Visual comparison of regional competency explanation methods for a speed limit dataset. Rows correspond to different input images and columns to different techniques. More red regions are more associated with low levels of model competency.}
    \label{fig:speed}
    \vspace{-10mm}
\end{figure}

The second dataset contains speed limit signs in Germany \cite{gtsdb}. While the training dataset is composed of common speed limit signs (30 km/hr and higher), the test dataset set also contains an uncommon speed limit (20 km/hr). This dataset assesses the ability of the regional competency approaches to identify regions associated with unseen classes. A visual comparison of these approaches is shown in Figure \ref{fig:speed} and a quantitative comparison is given in Table \ref{tab:speed}. We also consider the performance of existing class activation mapping techniques for identifying regions associated with unseen classes in Figure \ref{fig:cam-speed} and Table \ref{tab:cam-speed} of Appendix \ref{cams-speed}.

\begin{table}[h!]
\centering
\vspace{-2mm}
\caption{Quantitative comparison of regional competency explanation methods for a speed limit dataset. A true positive occurs when a method correctly identifies a pixel as one corresponding to a feature associated with an unseen class (i.e., the number 2), and a true negative occurs when a method identifies a pixel as one corresponding to a feature present in the training set (i.e., other parts of the traffic sign and background).}
\begin{tabular}{|c|c|c|c|c|c|c|}
\hline 
\multirow{2}{*}{Method} & \multirow{2}{*}{\shortstack{Average \\ Time (sec)}} & \multicolumn{5}{c|}{Average Accuracy (\%)} \\
\cline{3-7}
& & \hspace{2mm}Overall\hspace{2mm} & \hspace{2mm}TPR\hspace{2mm} & \hspace{2mm}TNR\hspace{2mm} & \hspace{2mm}PPV\hspace{2mm} & \hspace{2mm}NPV\hspace{2mm} \\
\hline 
Cropping & 24.27 & 22.21 & 54.37 & 21.08 & 2.30 & 93.06 \\
Masking & 5.17 & 47.57 & 20.32 & 48.44 & 4.55 & 92.14 \\
Perturbation & 5.16 & 83.76 & 32.24 & 85.51 & 12.77 & 97.33 \\
Gradients & 0.319 & 83.98 & \textbf{60.99} & 84.85 & 12.76 & 98.41 \\
Reconstruction & \textbf{0.109} & \textbf{92.19} & 58.14 & \textbf{93.43} & \textbf{26.20} & \textbf{98.44} \\
\hline
\end{tabular}
\vspace{-5mm}
\label{tab:speed}
\end{table}

Inspecting outputs across examples in the speed limit signs dataset, it appears that the \textit{Cropping} method is not useful for this dataset, offering very little information through its outputs. The \textit{Masking} method also does not appear very useful, as the outputs generally do not align with our expectations about the dependence of low model competency on elements of the unseen class (i.e., the number 2). Both of these methods have very low overall accuracy.

Compared to the first two methods, \textit{Perturbation} and \textit{Gradients} appear to offer more useful information. However, while both methods will sometimes assign higher dependency to the digit associated with the unseen class, the \textit{Perturbation} method is not the most reliable and the \textit{Gradients} method often places a higher dependency on the zero digit than it does on the two. While these two methods appear very similar in terms of their overall accuracy, the \textit{Gradients} method has a much higher TPR, while maintaining a similar TNR.

Both visually and quantitatively, the \textit{Reconstruction} method appears to perform the best. This method often indicates that low model competency is highly dependent on the digit 2, which is associated with an unseen class. It also achieves the highest overall accuracy, TNR, PPV, and NPV. In addition, this method has the lowest average computation time, maintaining a time of around 0.1 seconds.

\subsection{Dataset 3: Outdoor Park} \label{park}

\begin{figure}[h]
    \centering
    \includegraphics[width=\columnwidth]{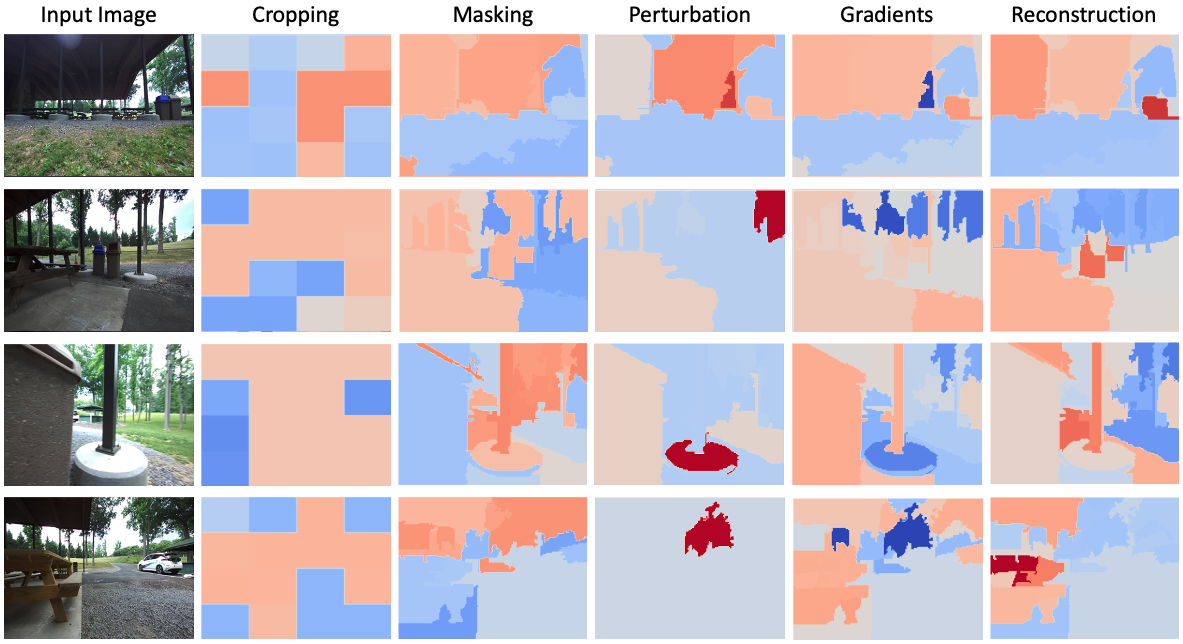}
    \caption{Visual comparison of regional competency explanation methods for an outdoor park. Rows correspond to different input images and columns to different techniques. More red regions are more associated with low levels of model competency.}
    \vspace{-5mm}
    \label{fig:pavilion}
\end{figure}

The third and final dataset contains images from regions in a park. While the training dataset only contains images from forested and grassy regions of the park, the test datset additionally includes images from around the park's pavilion. This dataset assesses the ability of the regional competency approaches to identify unexplored areas in an environment. A visual comparison of these approaches is shown in Figure \ref{fig:pavilion} and a quantitative one is given in Table \ref{tab:pavilion}. We also consider the performance of existing class activation mapping techniques for identifying unexplored regions in Figure \ref{fig:cam-park} and Table \ref{tab:cam-park} of Appendix \ref{cams-park}.

\vspace{5mm}

As with the speed limit signs dataset, the \textit{Cropping} method appears to offer little explanation for low model competency with the park dataset. The \textit{Masking} and \textit{Perturbation} methods sometimes identify unexplored regions as expected, but they do not consistently assign these regions higher incompetency dependency scores. These three methods all have low overall accuracy scores.

The \textit{Gradients} and \textit{Reconstruction} methods perform comparably, assigning high dependency scores to many of the image segments associated with unexplored areas and achieving similar levels of accuracy. Comparing these two methods, \textit{Gradients} achieves a higher overall accuracy, TPR, and NPV, but \textit{Reconstruction} achieves a higher TNR and PPV. The \textit{Reconstruction} method is faster than \textit{Gradients} in terms of computation time, but both methods are reasonably fast and could be used in most decision-making pipelines.

Because neither the \textit{Gradients} nor \textit{Reconstruction} method clearly outperforms the other, it is interesting to consider how they might be used in combination. The average accuracy achieved using the average of the outputs of these two methods is shown in the last row of Table \ref{tab:pavilion}. Combining these two methods increases the overall accuracy to over 80\% and the TPR to over 90\%. Using this technique, the NPV is also higher, but the TNR and PPV are lower. While using two methods would require more computational effort, the computation time is still within reason to be useful for decision-making.

\begin{table}[h]
\centering
\vspace{-2mm}
\caption{Quantitative comparison of regional competency methods for an outdoor park environment. A true positive occurs when a method correctly identifies a pixel as one corresponding to an unexplored region of the environment (i.e., areas around the pavilion), and a true negative occurs when a method identifies a pixel as one corresponding to a feature present in the training set (i.e., forested and grassy regions).}
\begin{tabular}{|c|c|c|c|c|c|c|}
\hline 
\multirow{2}{*}{Method} & \multirow{2}{*}{\shortstack{Average \\ Time (sec)}} & \multicolumn{5}{c|}{Average Accuracy (\%)} \\
\cline{3-7}
& & \hspace{2mm}Overall\hspace{2mm} & \hspace{2mm}TPR\hspace{2mm} & \hspace{2mm}TNR\hspace{2mm} & \hspace{2mm}PPV\hspace{2mm} & \hspace{2mm}NPV\hspace{2mm} \\
\hline 
Cropping & 0.261 & 37.82 & 48.48 & 17.20 & 53.51 & 15.73 \\
Masking & 0.301 & 34.51 & 29.00 & 43.60 & 44.73 & 24.73 \\
Perturbation & 0.330 & 37.67 & 22.47 & 67.14 & 27.76 & 30.48 \\
Gradients & 0.130 & 77.13 & 81.13 & 69.52 & 81.99 & 71.86 \\
Reconstruction & \textbf{0.053} & 75.56 & 79.35 & \textbf{73.81} & \textbf{84.53} & 67.49 \\
\hline 
Grad. + Reco. & 0.183 & \textbf{80.09} & \textbf{90.26} & 62.33 & 81.22 & \textbf{79.73} \\
\hline
\end{tabular}
\vspace{-5mm}
\label{tab:pavilion}
\end{table}

\subsection{Analysis of Results}

\begin{figure}[h]
    \centering
    \includegraphics[width=0.9\columnwidth]{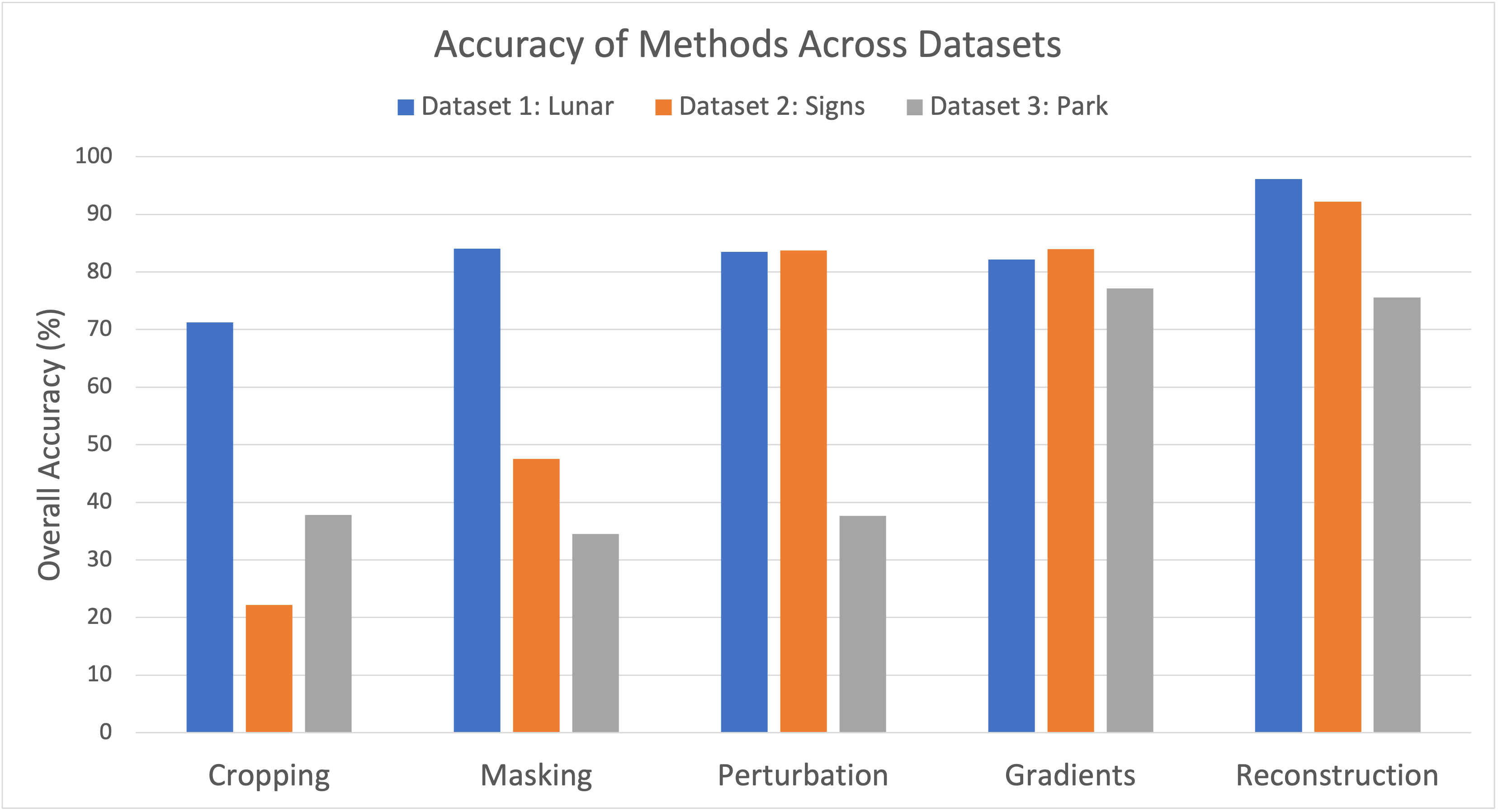}
    \caption{Comparison of the overall accuracy of the five regional competency explanation methods in identifying regions of an image that are unfamiliar to the perception model.}
    \label{fig:accuracy}
\end{figure}

Comparing all of our results across the three datasets in Figure \ref{fig:accuracy}, we see that the \textit{Cropping} method may provide a rough idea of image regions contributing to low model competency for simple datasets, but this method is unlikely to provide utility for more complex datasets. The \textit{Masking} and \textit{Perturbation} methods may be useful for some datasets, but it is unlikely that they would provide more explanatory power than the \textit{Gradients} and \textit{Reconstruction} methods. Among all of the regional competency methods, the \textit{Reconstruction} method appears the most promising for identifying regions contributing to low model competency, but the \textit{Gradients} method also has merit. While the \textit{Reconstruction} method more reliably informs us of regions in the image that are unfamiliar to the classifier, the \textit{Gradients} method more directly tells us which pixels the competency score actually depends on. Because these two methods inform us of slightly different aspects of model confidence, there may be value in combining their outputs.

\begin{figure}[h]
    \centering
    \includegraphics[width=0.9\columnwidth]{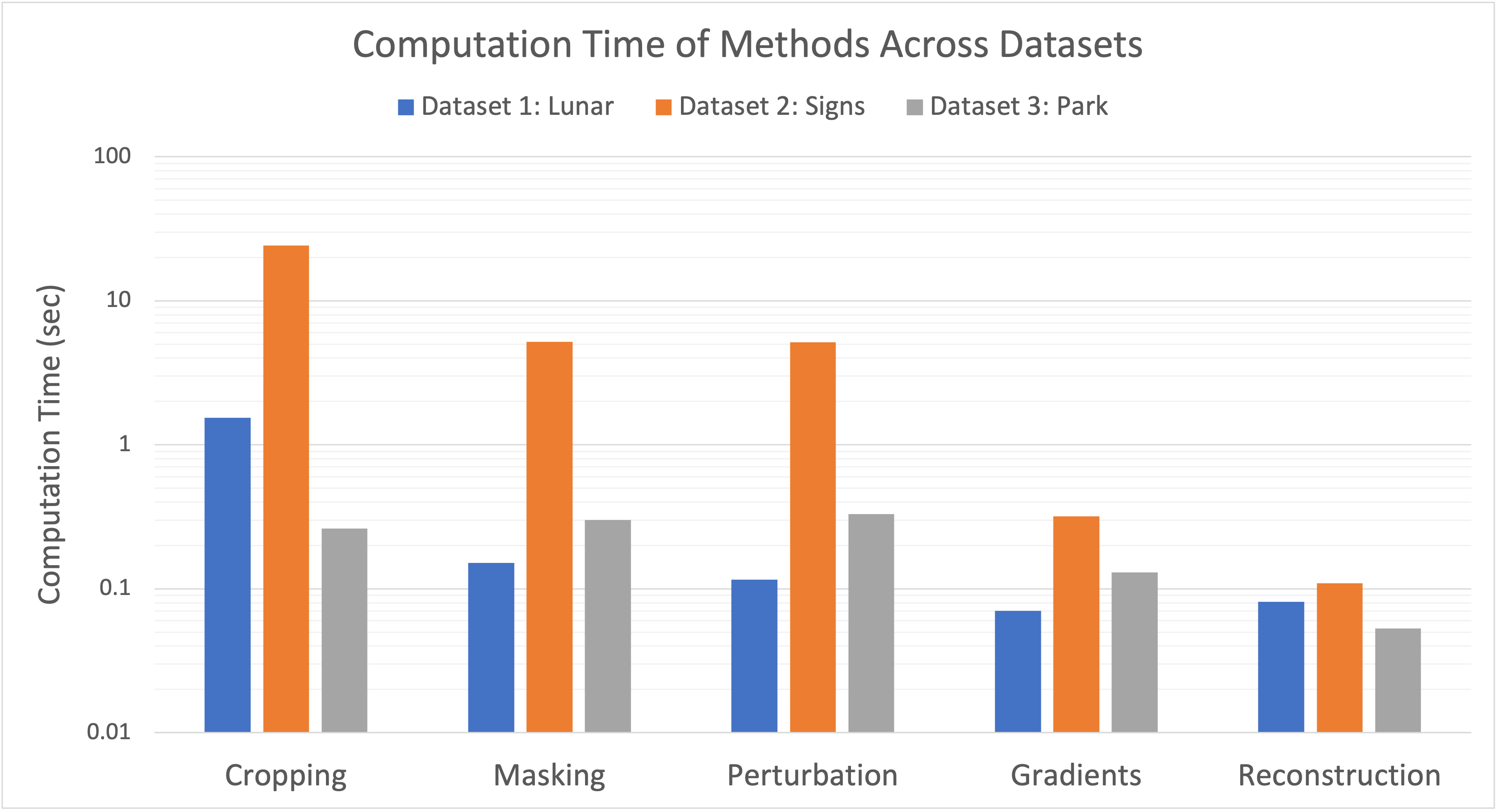}
    \caption{Comparison of the computation time of the five regional competency methods.}
    \label{fig:times}
\end{figure}

From Figure \ref{fig:times}, both the \textit{Gradients} and \textit{Reconstruction} methods can obtain their outputs in around 0.1 seconds or less on a 2.3 GHz Quad-Core Intel Core i7 processor. While these methods are fast enough to be usable in most decision-making pipelines, their computation times are dependent on image size, classifier architecture, and segmentation algorithm parameters, so computation time can vary across systems. While the increase in computation time seen for the speed limit signs dataset was not terribly significant for the \textit{Gradients} and \textit{Reconstruction} methods, it was quite significant for the \textit{Cropping}, \textit{Masking}, and \textit{Perturbation} methods. The \textit{Cropping} method is unlikely to run fast enough for most systems, and the \textit{Masking} and \textit{Perturbation} methods may be too slow for systems with larger images and a greater numbers of image segments.

From a visual comparison of the performance of the regional competency methods across the three datasets, it appears that these methods may be better at detecting unfamiliar objects and structures in an image (as with the lunar and park datasets), rather than identifying elements of unseen classes (as with the speed limit dataset). For the speed limit dataset, the `2' associated with the unseen class looks very similar to the `3' associated with a seen class. It is very possible that elements of the background appear more unfamiliar to the model than does the unseen digit. All five methods are much more precise when tasked with simply identifying features that are unfamiliar to the model, rather than those that the classifier should depend on but has not seen before. 

Furthermore, these methods are generally more successful at detecting objects and structures that do not look very similar to elements in the training set. The \textit{Gradients} and \textit{Reconstruction} methods were generally effective for the park dataset but tended to assign low dependency scores to the gravel in unexplored areas of the park, which is unfamiliar to the model but looks similar to dirt and short grass that the classifier has seen. While the \textit{Reconstruction} method was quite successful for the lunar dataset, the \textit{Gradients} method struggled to identify dark and shadowy objects that blended into the lunar background.

In addition, these methods were generally most successful for the lunar dataset, which is relatively simple and can be more clearly segmented into a few distinct regions. The Felzenszwalb segmentation algorithm \cite{felzenszwalb_efficient_2004} is useful because it is fast and does not need to be trained on a specific dataset to work effectively. However, it often produced strange results, particularly for the speed limit and park datasets, affecting the performance of the four approaches that utilized this algorithm. These methods may be improved by using a different segmentation algorithm--especially one that is domain-specific.

Overall, the \textit{Gradients} and \textit{Reconstruction} methods show promise for identifying regions associated with low model competency, particularly when aspects of the image that are unfamiliar to the perception model cause this reduction in competency. These methods are generally accurate and fast enough to be used for decision-making, but additional work may be done to improve these methods. 



\section{Conclusions}

In this paper, we sought to develop tools for understanding when DNN-based image classification models fail and why. We expand upon uncertainty quantification and OOD detection work to offer explanations for why a perception model lacks confidence in its prediction for a given input image by identifying specific regions in the image that lead to low model competency. Rather than offering explanations for classification decisions, as is typically done in explainable AI work, we aim to explain why a model is incompetent in its prediction.  

With these goals in mind, we developed five distinct methods for identifying regions in an image contributing to low model competency, which we refer to as (1) \textit{Image Cropping}, (2) \textit{Segment Masking}, (3) \textit{Pixel Perturbation}, (4) \textit{Competency Gradients}, and (5) \textit{Reconstruction Loss}. Each of these methods outputted an incompetency dependency score for each pixel in a set of input images. We evaluated each of these five approaches across three diverse datasets with the aim of determining which methods were most suitable for identifying unfamiliar objects, regions associated with unseen classes, and unexplored areas in an environment. We compared the dependency outputs of the five approaches visually, evaluated their ability to accurately identify image features that are unfamiliar to the classification model, and compared their computation times.

We found the \textit{Reconstruction Loss} method to be the most effective and promising approach overall but also found merit in the \textit{Competency Gradients} method. We believe both methods offer valuable information about why an image classification model is incompetent for a given input image, and their outputs can be obtained fast enough to be used in decision-making problems. 

\section{Limitations \& Future Work}

Additional work could be done to improve the performance of the regional competency estimation methods presented in this work. First, it may be worthwhile to explore different segmentation algorithms, particularly ones that are domain-specific or trained to work well with a particular dataset or environment of interest. For the \textit{Reconstruction} method, it would be interesting to consider additional reconstruction loss functions, as well as ensembles of autoencoders. While the \textit{Reconstruction} method is independent of the competency metric, the other four methods are directly dependent on the chosen metric. In this work, we only considered a single metric for competency, which incorporates different facets of uncertainty into a single score. It would be interesting to disentangle these sources of uncertainty or consider different metrics for competency. While the \textit{Cropping}, \textit{Masking}, and \textit{Perturbation} methods do not have strict requirements for the chosen competency metric, the \textit{Gradients} method is only useful if the competency score is differentiable with respect to the input image.

Beyond these more immediate next steps for improving the proposed methods, there are several other avenues for future work. First, this work focused on explaining low model competency for OOD images with regional features that are clearly unfamiliar to the perception model. Additional work would need to be done to understand low model competency for more nuanced cases, particularly with in-distribution images, and to understand high model competency. One could explore other methods for explaining model competency that are not focused on identifying particular regions contributing to low competency. It would be worthwhile to identify other types of image features that might contribute to low competency, such as image brightness. This work also seeks to understand perception model competency of trained models without modifying the underlying model architecture. It would interesting to also explore antehoc methods from the XAI literature to better understand model competency.

The approaches explored in this paper can be used to provide users a deeper understanding of how a model works and when it might fail. Additional work could be done with this goal in mind to combine visual explanations with language. Future work should also explore how the regional competency information presented in this paper should play into decision-making problems. While these methods provide perception-based systems with more information about why an image classification model is incompetent, it is still not clear how the system should respond. Additional work would need to be done to decide the appropriate response to model incompetency for different applications. Finally, the work presented in this paper focused on image classification problems. Additional work should be done to understand how these approaches could be adapted for regression, semantic segmentation, object detection, and other problems.

\subsubsection{\discintname}
The authors have no competing interests to declare that are
relevant to the content of this article.

\newpage
\appendix

\section{Data Sources \& Algorithm Parameters} \label{details}

To reproduce this work, data is available on \href{https://drive.google.com/drive/folders/120QyusyQvv1rXpcst9Wk0RI1RnpHn7Y_?usp=sharing}{Drive}, code is available on \href{https://github.com/sarapohland/explainable-competency}{GitHub}, and the chosen model architectures and training parameters are available on \href{https://drive.google.com/drive/folders/1kXVj7tXcLdV4TINsmARCflgy_GFm4xEU?usp=sharing}{Drive}. For more information on reproducing this work, please see this \href{https://github.com/sarapohland/explainable-competency/blob/main/README.md}{README}.

\section{Comparison to Class Activation Maps} \label{cams}

The goal of the work presented in this paper is to identify regions in an image contributing to low model competency and to display this incompetency dependency in a saliency map. This is closely related to work on developing class activation maps (CAMs), which are class-discriminative saliency maps that identify key regions in an image used by the model to make a particular class prediction. There are various methods for generating CAMs \cite{cam_survey}, but we choose to focus on a few representative approaches: Grad-CAM \cite{grad-cam}, Guided Grad-CAM \cite{grad-cam}, Grad-CAM++ \cite{grad-cam-pp}, Integrated Gradients \cite{Sundararajan}, SmoothGrad \cite{smooth-grad}, DeepLIFT \cite{Shrikumar}, Score-CAM \cite{score-cam}, Ablation-CAM \cite{ablation-cam}, Eigen-CAM \cite{eigen-cam}, and LayerCAM \cite{layer-cam}. These methods were implemented with the help of the PyTorch library for CAM methods \cite{jacobgilpytorchcam} and the Captum model interpretability library for PyTorch \cite{captum}.

The chosen mapping approaches are designed to explain model predictions, but we evaluate their ability to recognize unfamiliar objects (Section \ref{cams-lunar}), identify regions associated with unseen classes (Section \ref{cams-speed}), and determine unexplored regions in an environment (Section \ref{cams-park}). We find that these approaches are not suited for generating competency maps, as is done in our work.

\subsection{Dataset 1: Lunar Environment} \label{cams-lunar}

\begin{table}[h]
\centering
\vspace{-2mm}
\caption{Quantitative comparison of CAM methods for a lunar environment.}
\begin{tabular}{|c|c|c|c|c|c|c|}
\hline 
\multirow{2}{*}{Method} & \multirow{2}{*}{\shortstack{Average \\ Time (sec)}} & \multicolumn{5}{c|}{Average Accuracy (\%)} \\
\cline{3-7}
& & \hspace{2mm}Overall\hspace{2mm} & \hspace{2mm}TPR\hspace{2mm} & \hspace{2mm}TNR\hspace{2mm} & \hspace{2mm}PPV\hspace{2mm} & \hspace{2mm}NPV\hspace{2mm} \\
\hline 
Grad-CAM & 0.03 & 66.18 & 32.50 & 71.24 & 25.32 & 84.34 \\
Guided Grad-CAM & 0.04 & 76.24 & 30.97 & 83.88 & 35.94 & 88.89 \\
Grad-CAM++ & 0.04 & 60.80 & 35.76 & 65.88 & 22.19 & 85.76 \\
Integrated Gradients & 0.46 & 71.88 & 21.26 & 79.34 & 20.47 & 87.58 \\
SmoothGrad & 2.68 & 77.51 & 18.19 & 85.88 & 19.32 & 87.92 \\
DeepLIFT & 0.05 & 73.36 & 24.01 & 80.58 & 20.67 & 87.96 \\
Score-CAM & 0.13 & 62.40 & 38.94 & 67.26 & 23.42 & 87.18 \\
Ablation-CAM & 0.23 & 62.39 & 35.44 & 67.20 & 21.52 & 85.76 \\
Eigen-CAM & 16.20 & 71.90 & 36.22 & 78.21 & 31.17 & 89.05 \\
LayerCAM & 0.11 & 73.81 & 19.87 & 82.31 & 15.68 & 87.25 \\
\hline
\end{tabular}
\vspace{-5mm}
\label{tab:cam-lunar}
\end{table}

\newpage
\begin{figure}[h]
    \centering
    \includegraphics[width=\columnwidth]{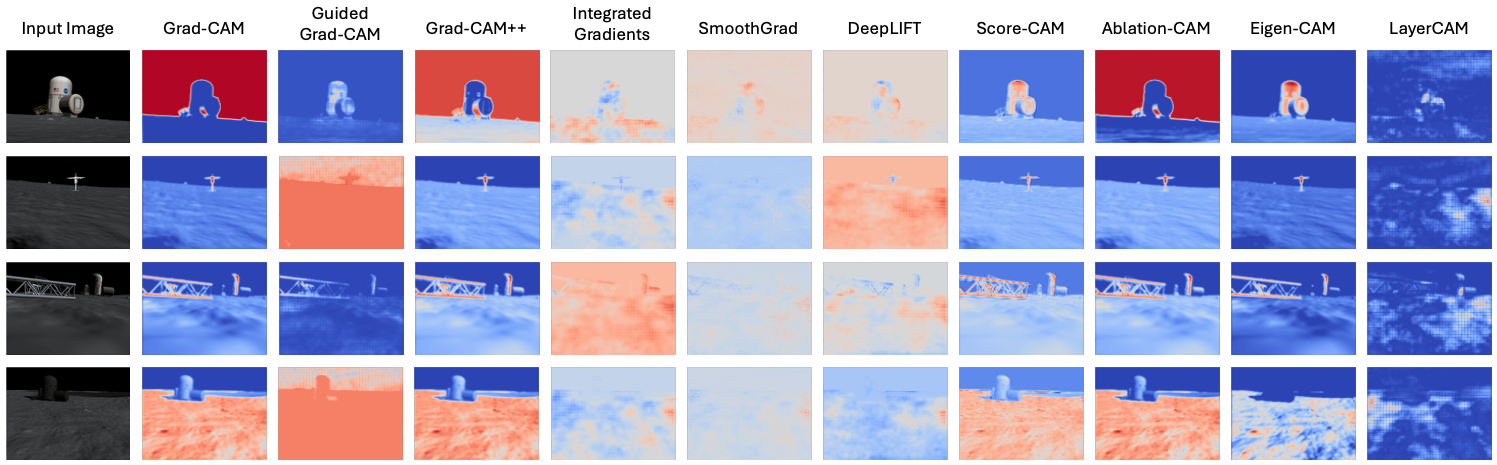}
    \caption{Visual comparison of class activation maps (CAMs) for a lunar environment. Rows correspond to different input images and columns to different techniques for generating saliency maps. More red regions contribute more to the model's prediction.}
    \label{fig:cam-lunar}
    \vspace{-10mm}
\end{figure}


\subsection{Dataset 2: Speed Limit Signs} \label{cams-speed}

\begin{table}[h]
\centering
\vspace{-2mm}
\caption{Quantitative comparison of CAM methods for a speed limit dataset.}
\begin{tabular}{|c|c|c|c|c|c|c|}
\hline 
\multirow{2}{*}{Method} & \multirow{2}{*}{\shortstack{Average \\ Time (sec)}} & \multicolumn{5}{c|}{Average Accuracy (\%)} \\
\cline{3-7}
& & \hspace{2mm}Overall\hspace{2mm} & \hspace{2mm}TPR\hspace{2mm} & \hspace{2mm}TNR\hspace{2mm} & \hspace{2mm}PPV\hspace{2mm} & \hspace{2mm}NPV\hspace{2mm} \\
\hline 
Grad-CAM & 0.41 & 69.88 & 36.78 & 71.01 & 4.53 & 97.07 \\
Guided Grad-CAM & 0.60 & 83.15 & 17.39 & 85.39 & 5.65 & 96.59 \\
Grad-CAM++ & 0.41 & 51.01 & 18.57 & 52.03 & 1.71 & 94.87 \\
Integrated Gradients & 18.06 & 51.43 & 49.46 & 51.50 & 3.33 & 96.78 \\
SmoothGrad & 80.19 & 50.30 & 49.89 & 50.31 & 3.28 & 96.78 \\
DeepLIFT & 0.87 & 52.90 & 48.20 & 53.06 & 3.35 & 96.80 \\
Score-CAM & 4.18 & -- & -- & -- & -- & -- \\
Ablation-CAM & 4.11 & 51.72 & 25.95 & 52.54 & 2.17 & 95.30 \\
Eigen-CAM & 88.97 & 71.66 & 62.17 & 71.99 & 6.78 & 98.27 \\
LayerCAM & 0.62 & 87.59 & 54.00 & 88.78 & 14.09 & 98.21 \\
\hline
\end{tabular}
\vspace{-5mm}
\label{tab:cam-speed}
\end{table}

\begin{figure}[h!]
    \centering
    \includegraphics[width=\columnwidth]{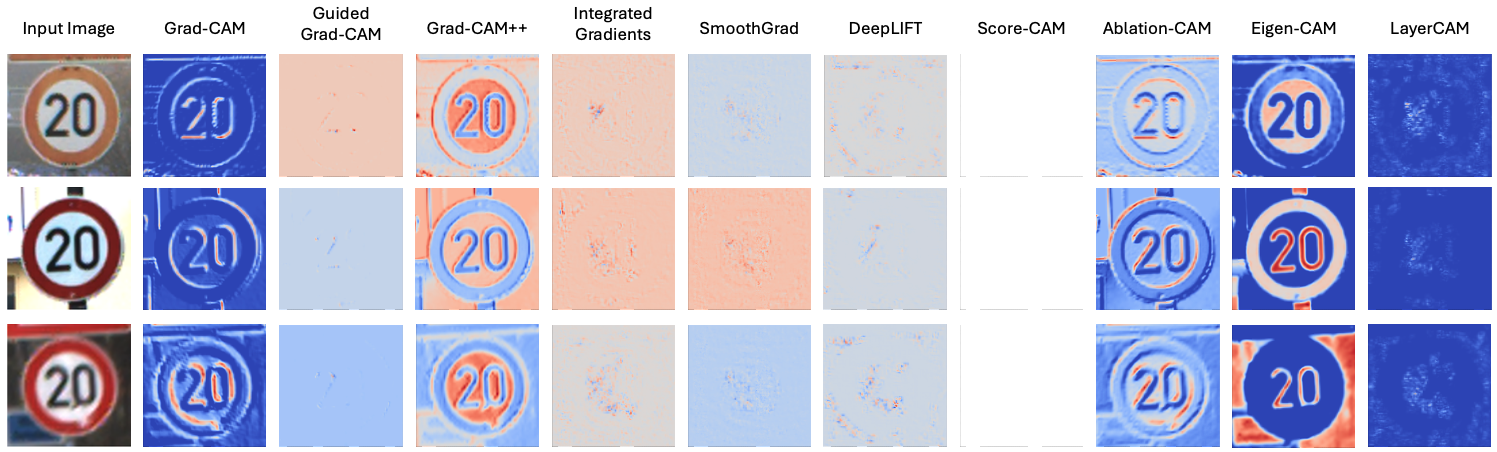}
    \caption{Visual comparison of class activation maps (CAMs) for a speed limit dataset. Rows correspond to different input images and columns to different techniques for generating saliency maps. More red regions contribute more to the model's prediction.}
    \label{fig:cam-speed}
\end{figure}


\subsection{Dataset 3: Outdoor Park} \label{cams-park}

\begin{table}[h]
\centering
\vspace{-2mm}
\caption{Quantitative comparison of CAM methods for an outdoor park environment.}
\begin{tabular}{|c|c|c|c|c|c|c|}
\hline 
\multirow{2}{*}{Method} & \multirow{2}{*}{\shortstack{Average \\ Time (sec)}} & \multicolumn{5}{c|}{Average Accuracy (\%)} \\
\cline{3-7}
& & \hspace{2mm}Overall\hspace{2mm} & \hspace{2mm}TPR\hspace{2mm} & \hspace{2mm}TNR\hspace{2mm} & \hspace{2mm}PPV\hspace{2mm} & \hspace{2mm}NPV\hspace{2mm} \\
\hline 
Grad-CAM & 0.03 & 46.62 & 33.22 & 74.19 & 72.27 & 35.69 \\
Guided Grad-CAM & 0.04 & 34.43 & 13.30 & 78.22 & 60.28 & 30.69 \\
Grad-CAM++ & 0.04 & 30.60 & 26.10 & 49.83 & 45.23 & 22.01 \\
Integrated Gradients & 0.46 & 30.69 & 13.25 & 67.67 & 34.40 & 28.66 \\
SmoothGrad & 2.65 & 33.98 & 22.23 & 59.38 & 43.78 & 28.68 \\
DeepLIFT & 0.05 & 31.57 & 18.97 & 57.25 & 45.10 & 26.82 \\
Score-CAM & 0.13 & 29.05 & 21.72 & 42.66 & 40.99 & 22.07 \\
Ablation-CAM & 0.22 & 32.16 & 20.99 & 55.19 & 46.16 & 25.71 \\
Eigen-CAM & 15.79 & 27.22 & 16.82 & 47.69 & 36.88 & 23.51 \\
LayerCAM & 0.10 & 31.27 & 14.96 & 63.28 & 44.79 & 27.66 \\
\hline
\end{tabular}
\vspace{-5mm}
\label{tab:cam-park}
\end{table}

\begin{figure}[h]
    \centering
    \includegraphics[width=\columnwidth]{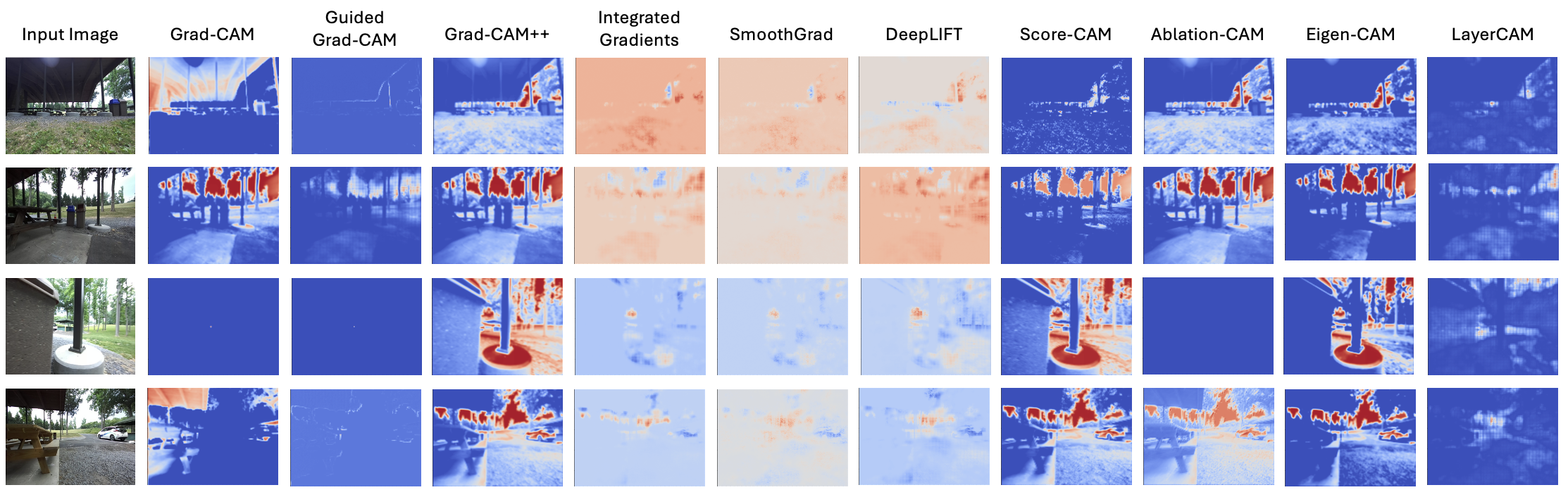}
    \caption{Visual comparison of class activation maps (CAMs) for an outdoor park. Rows correspond to different input images and columns to different techniques for generating saliency maps. More red regions contribute more to the model's prediction.}
    \label{fig:cam-park}
\end{figure}


\newpage
\bibliographystyle{IEEEtran}
\bibliography{citations}

\end{document}